\documentclass[letterpaper, 10 pt, conference]{ieeeconf}  
\IEEEoverridecommandlockouts                              

\usepackage{graphicx}
\usepackage{array, makecell} 
\graphicspath{ {./img/} }

\title{\LARGE \bf
Labeler-hot Detection of EEG Epileptic Transients
}

\author{Lukasz Czekaj$^{1}$, Wojciech Ziembla$^{1}$, Pawel Jezierski$^{2}$, Pawel Swiniarski$^{1}$, \\Anna Kolodziejak$^{1}$, Pawel Ogniewski$^{1}$, Pawel Niedbalski$^{1}$, Anna Jezierska$^{3,4}$, Daniel Wesierski$^{4}$
\thanks{$^{1}$Elmiko Biosignals, Poland}%
\thanks{$^{2}$Institute of Psychiatry and Neurology, Poland}%
\thanks{$^{3}$Systems Research Institute of the Polish Academy of Sciences, Poland}%
\thanks{$^{4}$Gdansk University of Technology, Poland}%
}

\begin{document}

\maketitle
\thispagestyle{empty}
\pagestyle{empty}

\begin{abstract}
Preventing early progression of epilepsy and so the severity of seizures requires an effective diagnosis. Epileptic transients indicate the ability to develop seizures but humans overlook such brief events in an electroencephalogram (EEG) what compromises patient treatment. Traditionally, training of the EEG event detection algorithms has relied on ground truth labels, obtained from the consensus of the majority of labelers. In this work, we go beyond labeler consensus on EEG data. Our event descriptor integrates EEG signal features with one-hot encoded labeler category that is a key to improved generalization performance. Notably, boosted decision trees take advantage of singly-labeled but more varied training sets. Our quantitative experiments show the proposed labeler-hot epileptic event detector consistently outperforms a consensus-trained detector and maintains confidence bounds of the detection. The results on our infant EEG recordings suggest datasets can gain higher event variety faster and thus better performance by shifting available human effort from consensus-oriented to separate labeling when labels include both, the event and the labeler category.
\end{abstract}

\section{INTRODUCTION}
Misinterpretation of scalp electroencephalogram (sEEG) is not uncommon in clinical practice \cite{bagheri2017interictal},\cite{dericioglu2018success}. At the same time, it can have severe negative consequences on health and well-being of patients undergoing epileptic diagnosis \cite{halford2017characteristics}. Developing algorithms that reliably assist clinicians in EEG inspection is thus an important challenge. 

Epilepsy is a chronic disease that affects dozens of millions of people worldwide, being the second neurological disorder after stroke. Nearly 85\% of the affected population belongs to developing countries. Roughly 2.4 million new cases of epilepsy occur every year globally. Epilepsy is often a consequence of motor vehicle accidents. As its occurrence increases with age, aging societies are especially at risk to suffering from epilepsy. Patients with epilepsy have a mortality rate significantly higher than that of the general population \cite{lhatoo2005cause}. 


Meanwhile, diagnostics of the disease can be time-consuming -- from hours to days, is expensive, and requires long clinical experience of the personnel. Gold-standard procedure for diagnosing epilepsy is measuring the electric activity of the cortex with sEEG. The modality uses a lattice of electrodes that are placed along the scalp. Inspection of EEG aims at finding patterns that mark abnormal electric activity of the brain. Among them, transient epileptic patterns indicating tendency toward seizures are of special interest. 

The prevalent approach to detection of epileptiform EEG discharges relies on machine learning algorithms that train a decision function in some feature space on an annotated dataset of EEG micro events. Recent validation studies show that human experts continue to outperform algorithms in detection of epileptiform discharges in sEEG \cite{halford2018interictal}. However, annotating pathological events, such as spikes, sharp waves, slow waves, and their complexes, is far from evident. A human expert can confuse pathological with benign events as they can share similar morphology \cite{santoshkumar2009prevalence}. Low signal-to-noise ratio and the presence of artifacts are other confounding causes of labeling errors \cite{halford2009computerized}. 

\begin{figure}
    \includegraphics[width=\linewidth]{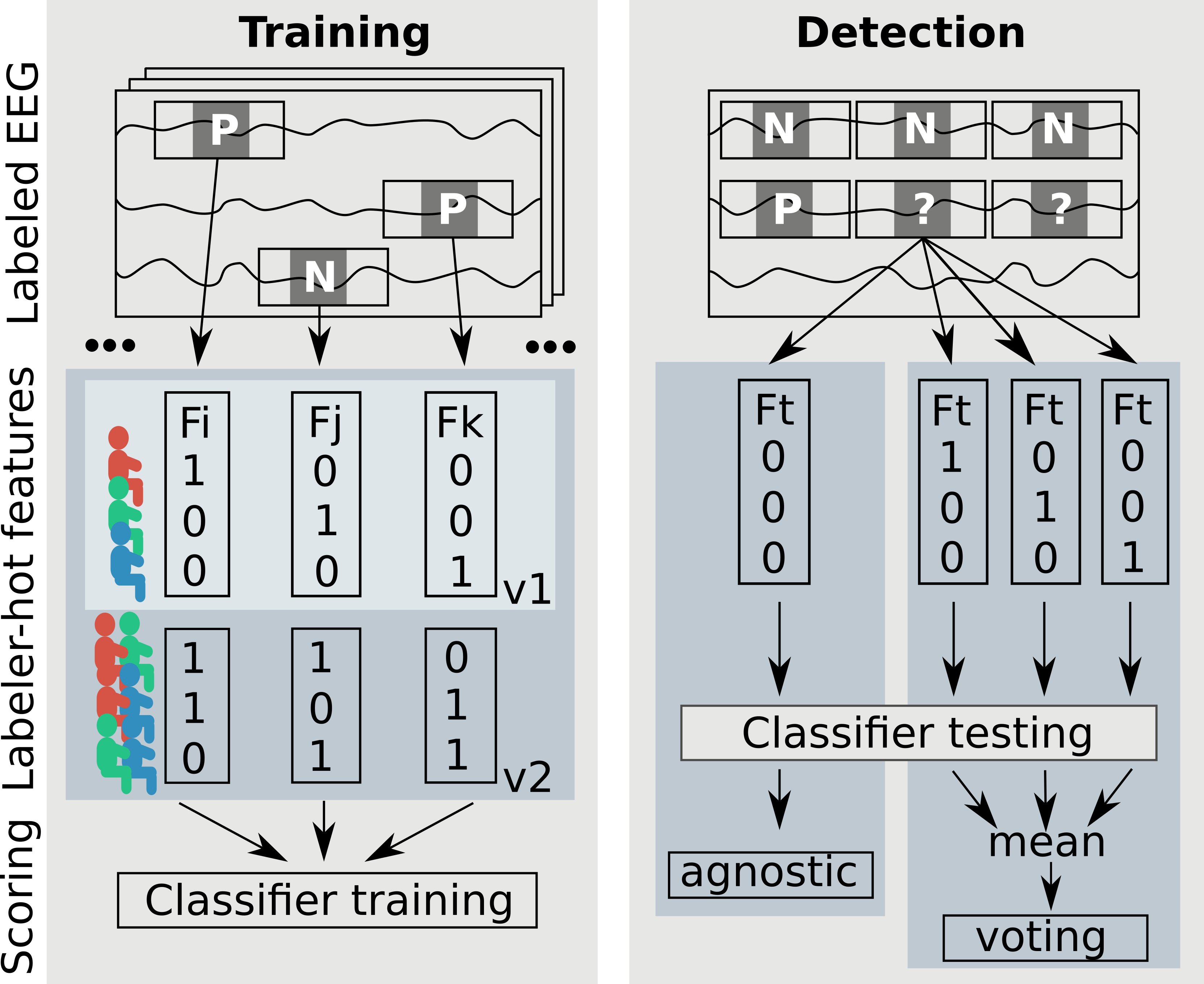}
    \caption{{\it Labeler-hot detection of EEG events.} In the training phase, a single labeler (v1) or a subgroup of labelers (v2) is appended into the F-descriptor of EEG event examples (here {\it i,j,k}) through one-hot encoding. Binary, boosted decision trees classifier learns to separate Pos/Neg events in the joint labeler-signal space. As the detection phase provides no information about the labeler, we either zero-pad the descriptor of each tested event (labeler agnostic case) or form labeler-specific descriptors followed by averaging their classification scores (labeler voting case).}
    \label{fig:method}
\end{figure}

Datasets are annotated, in effect, by a designated group of hospital personnel, with multiple but noisy labels per event. Ground-truth event labels for training and testing usually are obtained then through majority-voted consensus of labelers \cite{bagheri2017interictal},\cite{scheuer2017spike},\cite{halford2013standardized}. However, the amassed multiple labels show only low-to-moderate inter-rater agreement (IRA) \cite{halford2013standardized}. The majority of neurologists have no neurophysiology fellowship training and there is a substantial discrepancy in event interpretations between board-certified academic clinical neurophysiologists \cite{bagheri2017interictal}. Moreover, technicians, who are more available than clinicians for annotating EEG \cite{scheuer2017spike}, often have less clinical experience and qualifications. The features of raters were analyzed in \cite{halford2017characteristics} that generally concluded the highest IRA was attributed to board-certified annotators. The groups of features of EEG signal, in turn, were selected and evaluated in \cite{bagheri2017interictal} that indicated wavelets led to higher IRA.

This work addresses the problem of training an EEG event detector on single and multiple labels per event when labels are provided by imperfect experts. Traditional scenario for training an EEG event detection algorithm has relied on consensus of the majority of labelers that determined ground truth event labels. The multiple labels have only low-to-moderate IRA though. We go beyond labeler consensus on EEG data. We demonstrate that a detector can improve its recall-precision performance noticeably through training on singly instead of multiply labeled events when more events are sampled across time and recordings, thereby increasing variety of training data, provided that the event descriptor identifies the event labeler. We achieve this by integrating groups of signal features with one-hot encoded labeler category in boosted decision trees training regime. The classifier then selects optimal feature subsets of epileptic events for training the event detector. We show that the proposed labeler-hot features are a key to higher generalization performance of the classifier. To our knowledge, we are the first to train EEG classifiers from consensus-free labels of imperfect experts.



\section{RELATED WORK}


Detection of epileptiform EEG discharges has a long tradition in EEG analysis. For comprehensive review see \cite{spikeDetectionReview2002, spikeDetectionReview2018}. In this section we describe methods that relate to our problem of training classifiers from noisy labels.

Ground truth can be estimated from multiple, noisy labels using crowdsourcing. Besides naive majority voting, more sophisticated algorithms, based e.g. on EM and labeler reliability estimation, were proposed in \cite{DawidSnake1979, Zhang2014, rodrigues2014gaussian} but require high redundancy of labels \cite{BraggWeld2016}. Recently, to overcome high redundancy constraint, an EM algorithm used predicated label as ground truth to estimate labeler confusion matrix \cite{Khetan2018}. There are also results specifically in the area of time series labeling, which are more related to EEG annotations than image labeling \cite{Gupta2016ModelingMT,Wang2018TowardsBetterGoldStandard}.

Another line of works tweaks loss function to incorporate assumption about uniform noise process disturbing labels \cite{Natarajan2013, Khetan2018, Gupta2016ModelingMT}. There was significant amount of work in the area of active learning \cite{Whitehill2009, Khetan2016} that ask for more labels of inconsistent examples. Allocation of work (i.e. multiple labeling vs single labeled but larger dataset) was studied in \cite{Sheng2008, Ipeirotis2014} finding that repeated labeling performs better if quality of labelers is below some threshold.

Unlike other approaches, that model labeler quality weights from training examples, we focus on modeling individual labeler "styles". Specifically, our approach attempts to predict which labeler says what about given EEG example. To our knowledge, similar approaches were used for the first time in \cite{guan2018said}, then generalized as "crowd layer" in \cite{Rodrigues2018crowds}, and for time series annotations in \cite{Huang2018crowdLayer}. Approach presented in \cite{guan2018said} is based on learning a logistic regression classifier for each labeler on the features obtained from Inception-V3 network. Then single labeler scores are aggregated by weighted averaging. In the training phase, loss function takes into account only the output corresponding to the labeler who provided the example. 


\begin{figure}
    \includegraphics[width=\linewidth]{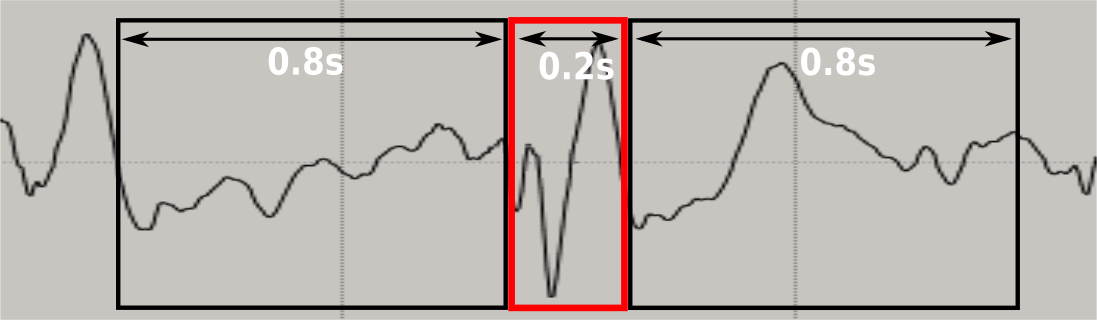}
    \caption{{\it Signal description} - an EEG fragment is converted into an array of descriptive parameters (sec. \ref{sec:descriptor}) that are computed within its central window (red box) and its left and right adjacent windows (black boxes).}
    \label{fig:descriptor_explanation}
\end{figure}

Our approach uses XGBoost learning and explicitly, one-hot encoded labeler as a feature. That differs our approach from \cite{guan2018said}. We also evaluate different methods of detection at test time as then no labeler information is provided. Moreover, we demonstrate our approach on EEG time series annotations instead of image labeling. 








\section{METHOD}


We address the task of detecting epileptic EEG micro-events in a single channel. Our detector processes each channel regardless of other channels. The flowchart of our method is depicted in Fig. \ref{fig:method}.

\subsection{Signal description}
\label{sec:descriptor}


Our descriptor is composed of three windows, a central window of $0.2$ sec. duration and two neighbourhood windows of $0.8$ sec. duration each (Fig. \ref{fig:descriptor_explanation}). Then, we calculate the following features of the windowed EEG signal and stack them column-wise into a descriptor:

\begin{itemize}
    \item Time series anomaly score, i.e. linear model prediction error,
    \item FFT features (log power for frequency) in central window, neighbourhood and quotient of window and neighbourhood features,
    \item Teager Energy for central window, neighbourhood and their quotient,
    \item Quotient of waveform length for window and neighbourhood,
    \item Standardised statistics (mean, standard deviation, skewness, min, max) of continuous wavelet (Ricker wavelet) transform coefficients for central window; we use signal standardization according to the neighbourhood,
    \item Statistics (mean, standard deviation, skewness, min, max) of EEG signal difference with lag $1$ in the central window.
\end{itemize}

\subsection{Learning}
\label{sec:learning}
The experience of labelers manifests itself in specific expert annotation styles that can be learned by the event detection algorithms. To this end, we propose to integrate the signal descriptor with the one-hot encoded labeler category. We explore $2$ variants of such an encoding: 
\begin{enumerate}
    \item single expert category (v1) -- each expert corresponds to one row in the descriptor; we set to $1$ the row of the expert who annotated a given example, otherwise the row is $0$,
    \item pair of experts category (v2) through one-hot encoding of single experts and one-hot encoding of 2-expert groups -- the same rows as in v1 and one row for every combination of a pair of experts; we set to $1$ the rows that correspond to groups that contain the expert who annotated a given example, otherwise the row is $0$.
\end{enumerate}

We then use an XGBoost classifier for training our decision function. The features that describe the signal and the labelers are input together with a class label to the XGBoost classifier. The classifier outputs predictions as probability that a tested example is positive. 

\subsection{Detection}
\label{sec:detection}

During training, we have information about who labeled what but for test examples we lack such cues. Hence, we propose two detection methods:
\begin{enumerate}
    \item expert agnostic -- all rows of descriptor related to labelers are set to $0$, what neglects labeler style,
    \item expert voting -- each example is considered to have been individually annotated by each expert and thus is processed as a set of (v1) descriptors. Then, the prediction outputs are mean-averaged as a final result.   
\end{enumerate}



\section{EXPERIMENTS AND RESULTS}

\subsection{EEG dataset of infants with tuberous sclerosis}
Our dataset (Tab. \ref{tab:dataset_summary}) consists of $30$ EEG recordings (sampling rate $256$ Hz) of infants with tuberous sclerosis. The dataset is split into $24$ training recordings from 18 patients and $6$ test recordings from 6 patients. Inpatient age span is 3-14 months and 2-26 months in the training and test recordings, respectively. 

Each $\sim1$h long recording of $18$ EEG channels, configured in bipolar Banan 2 montage (20-10 standard), is annotated within $5$ blocks that come from various locations in the recording. Each block is 5 sec. long. A group of experienced EEG technicians individually annotates the same blocks by placing adjacent event windows of variable duration along each channel. As the labelers can decide when a given event starts and ends on the time axis, our annotation protocol limits the allowable duration of the windows to $2$ sec. in order to encourage the labelers to look at local EEG fragments along each channel. Each event window is categorized either as: (N-negative) artifact, slow wave, sleep spindle, norm, other, (P-positive) sharp wave, spike, sharp wave and spike complexes. All recordings and annotations were acquired with Elmiko EEGDigiTrack hardware and software.

\begin{table}
    \centering
    \begin{tabular}{l|c|r|r|c|r|r}
    &\multicolumn{3}{c|}{training data}&\multicolumn{3}{c}{test data}\\
labeler&recording&\#pos&\#neg&recording&\#pos&\#neg\\\hline\hline
L1&R7-30&3118&542K&R1-6&1857&729K\\\hline
L2&R7-30&498&546K&R1,3-6&374&542K\\\hline
L3&
\makecell{R8-14,16-21,\\23,26-28}
&1199&380K&R1,3-6&981&545K\\\hline
L4&--&--&--&R1,2&270&357K\\\hline
L5&--&--&--&R2&342&196K\\\hline
L6&
\makecell{
R7,15,22,24,\\
25,29,30}
&288&163K&R2&347&195K\\\hline
L7&--&--&--&R4&616&90K\\
    \end{tabular}
    \caption{Summary of our EEG dataset of epileptic transients in infants with tuberous sclerosis (K$=\times10^3$). 
    }
    \label{tab:dataset_summary}
\end{table}

\begin{figure}
    \includegraphics[width=\linewidth]{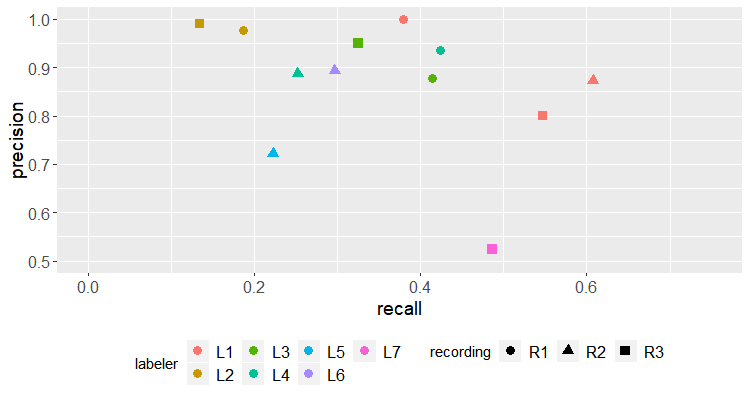}
    \caption{{\it Labeler quality} - we show precision and recall scores of each labeler. Each labeler is compared against the ground truth, obtained from majority voting of $3$ other labelers. Results are presented for $3$ test recordings.}
    \label{fig:prec_rec_labeler}
\end{figure}

\subsection{Deriving consensus labels and cropping event windows}
\label{sec:consensuslabels}
The consensus-based label per event was determined by the dominant class within the labels that were assigned to the event by $3$ experts. As experts annotated EEG channels independently, their event windows start and end at different time locations on the given EEG channel. To account for this misalignment, the overlap of two events of the same class produced the cropped window of the event. Then, for the positive class, the center of that window is the center of the central window of the signal descriptor (Fig. \ref{fig:descriptor_explanation}). As the misaligned negative class windows can reach 2 sec., their overlapping window is usually longer than $0.2$s. Our procedure then produces negative event centers every 0.1 second along the cropped window. 

\subsection{Test data and evaluation protocol}
The ground truth for the test set was obtained through the consensus of 3 experts (sec. \ref{sec:consensuslabels}). We prepare $5$ sets of negative test examples by randomly sampling the whole negative testset. The positive testset is fixed and has $4223$ events. Each pair of generated negative and positive testsets is imbalanced, where the count of negative to positive examples is $20:1$ thereby reflecting up to some degree the inherent prevalence of negative events in an EEG recording.

\textbf{Evaluation metric} We used average precision (AP) score from the precision-recall curve to evaluate our method. If more than $3$ experts labeled a recording, combinations of 3 expert labels give rise to multiple ground-truth labels per event. In this case, we average the AP scores for each such recordings. The final AP score is the average over the AP scores of the recordings. In this way, only the number of recordings, and not the number of labelers per recording, affects the final score.

\textbf{Labelling quality} In order to gain insight into the annotating quality of the experts, we compare their individual performance with respect to the consensus-based ground truth from the rest of labelers on the test data in Fig. \ref{fig:prec_rec_labeler}. The labelers tend to have higher precision than recall indicating that experts can miss epileptic events in the electroencephalogram.

\subsection{Training data: scenarios for learning from noisy labels}
\label{sec:scenarios}

We describe four scenarios for designating imperfect experts to annotating micro-events in EEG recordings. Notably, assuming budget, availability, and time constraints, we ask whether a group of medical labelers should annotate (i) the same recordings in the same time instants (A,B), (ii) the same recordings but at different time instants (C), or (iii) different recordings (D) in order to build a dataset that will train the best performing event detector.

In the A-scenario, ground truth labels are consolidated based on the consensus of $K=3$ labelers (sec. \ref{sec:consensuslabels}). In the B-scenario, the raw EEG examples are the same as in the A-scenario. However, as an event was labeled $K$-times by individual labelers, it can belong to opposite training sets in the B-scenario. The A-scenario has $K$-times fewer training data than do the B,C,D-scenarios, which have single labels per event. Importantly though, the group of experts perform the same amount of work in each scenario.

\begin{figure}[!htbp]
    \includegraphics[width=\linewidth]{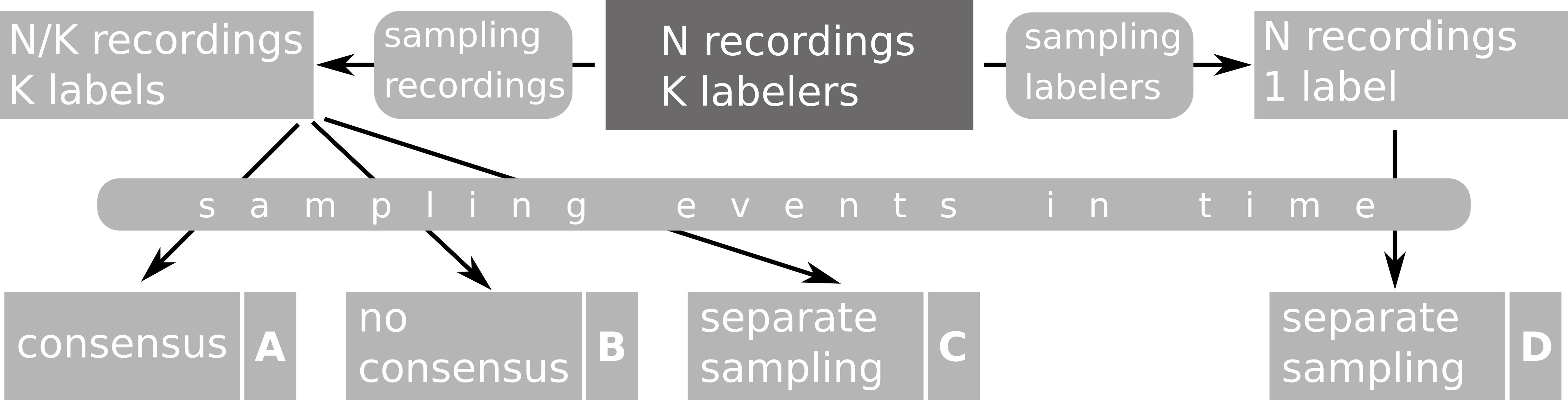}
    \caption{{\it Scenarios A--D for training EEG event detectors from noisy labels.} }
    \label{fig:experiment_schema}
\end{figure}

\textbf{Sampling} We are given $N=24$ recordings in the training dataset, multiply annotated by $K=3$ labelers (see Fig. \ref{fig:experiment_schema}). We randomly sample $5$ times either (i) $8$ out of $24$ recordings such that $8$ recordings have $3$ labelers (A,B,C) or (ii) a disjoint assignment of $3$ labelers to $24$ recordings such that $8$ recordings have $1$ labeler (D). Then, we sample $5$ times $100$ positive and $100$ negative examples on the time axis per each sampled recording. In total, we have $25$ different realisations of training data in the form of (recording, labeler)-pairs in each scenario with $2\cdot800$ and $2\cdot3\cdot800$ examples for A and B,C,D scenarios, respectively.

\subsection{Quantitative results}

The agnostic-based and voting-based detector (sec. \ref{sec:detection}), that used either v1 or v2 descriptors (sec. \ref{sec:learning}) and was trained under the C-scenario, performed better than the detector, that was trained under the A-scenario. Although the voting-based detector performed slightly better than the agnostic-based detector by median AP score of $0.2-0.5\%$, we use the agnostic detection method in the remaining experiments.  

\begin{figure}[!htbp]
    \includegraphics[width=\linewidth]{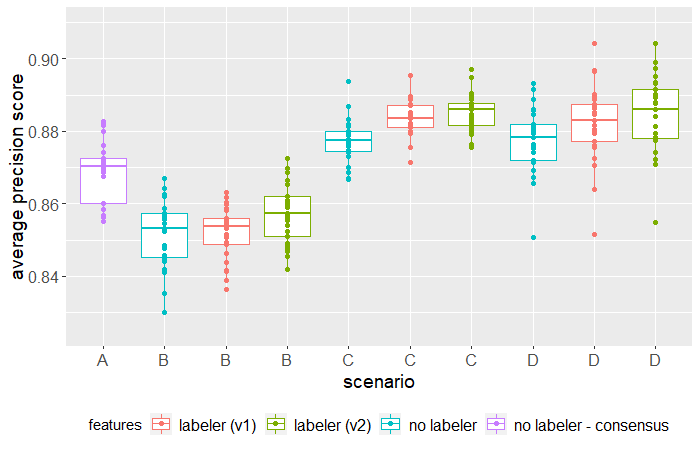}
    \caption{{\it Comparison of learning scenarios} - box plots of average precision scores for scenarios A,B,C,D (sec. \ref{sec:scenarios}), descriptors v1,v2 (\ref{sec:learning}), and agnostic detection method (\ref{sec:detection}). Each point represents a single experiment. The B-scenario has poorest performance. We observe including labeler category into the signal descriptor (C,D-scenarios) leads to systematically better results and maintains the confidence bounds of the detection wrt to consensus-based detector (A-scenario).}
    \label{fig:scenario_comparison}
\end{figure}

\begin{figure}[!htbp]
    \includegraphics[width=\linewidth]{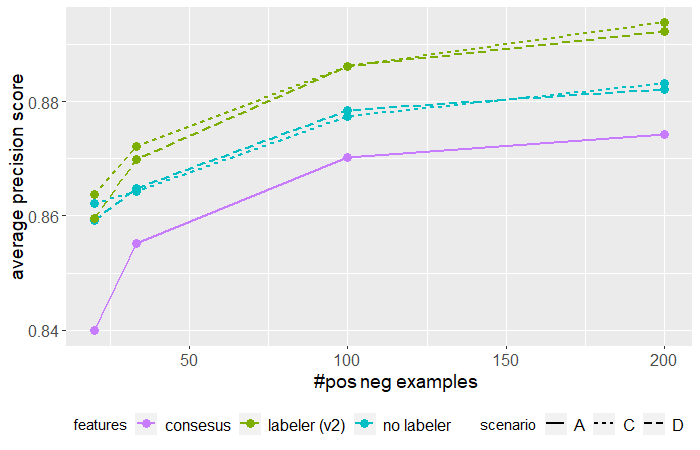}
    \caption{{\it Event detection performance wrt the increasing number of training examples per recording.} We present median AP scores of the detectors. Detectors, trained on singly labeled, higher event variability datasets (C,D) perform $\sim 1\%$ better than detectors, trained on consensus-based datasets (A). Detectors, trained under C,D-scenarios and with labeler-hot descriptors (v2), push the performance curves by another $\sim 1\%$ higher.}
    \label{fig:samp_cnt_dep}
\end{figure}

We evaluate the scenarios from sec. \ref{sec:scenarios} in Fig. \ref{fig:scenario_comparison}. We find that training event detectors on datasets created by allocating labelers to disjoint annotations on time axis (C) and to different recordings (D) improves median AP score by $\sim1.5-2\%$ over datasets created by consensus-based annotations (A). Including the labeler category (v1,v2) into the event descriptor helps in every scenario (B,C,D). The B-scenario is always worse than the A-scenario. We posit this is due to the fact that the B-scenario produces the same dataset variability as the A-scenario but introduces conflicting labels per event. Collectively, the results indicate that detectors achieve best performance by increasing the variety of training data and at the same time by including the labeler category into event description. These remarks are further emphasized in Fig. \ref{fig:samp_cnt_dep} by increasing the volume of training datasets from the A,C,D-scenarios.

\subsection{Implementation details}

For scenarios A-D, we trained the XGBoost decision trees with binary logistic loss and with all possible configurations of training parameters. We show only the best performing classifiers in each scenario. The configuration parameters are: max tree depth $\{5, 10\}$, learning rate $\{0.005, 0.01\}$, column subsampling $\{0.1, 0.2, 0.5\}$, row subsampling $\{0.5\}$, number of trees $\{1000, 2000\}$. Feature extraction for $2\cdot2400$ training examples (B,C,D scenarios) and $21\cdot4223$ testing examples takes $\sim2$hrs. For each sampled train/test set and for particular training parameter configuration, training the trees and evaluating them jointly takes $\sim20$min. All experiments were implemented in Python $3.6$ and were run on a PC with $64$ GB RAM and CPU Intel i7 $3.4$GHz. 

\section{CONCLUSIONS}
We describe an effective approach to leveraging individual expertise of medical labelers. Experts have unique strengths in annotating specific EEG data -- some experts might feel more comfortable with annotating artifacts while others with annotating spikes. Such expert preferences manifest themselves in specific annotation styles that can be learned by the event detection algorithms. To this end, our approach integrates the signal descriptor with variants of one-hot encoded labeler categories and shifts available human effort from consensus-oriented to separate labeling thereby increasing the variety of the training dataset. Enhancing the training procedure jointly by both propositions is a key to increased performance of EEG event detectors at test time.


\section*{ACKNOWLEDGMENT}

This work was financed in part by the Epimarker project under agreement STRATEGMED3/306306/4/NCBR/2017.

\bibliographystyle{unsrt}
\bibliography{eusipco2019_eeg}

\begin{thebibliography}{10}

\bibitem{bagheri2017interictal}
Elham Bagheri, Justin Dauwels, Brian~C Dean, Chad~G Waters, M~Brandon Westover,
  and Jonathan~J Halford.
\newblock Interictal epileptiform discharge characteristics underlying expert
  interrater agreement.
\newblock {\em Clinical Neurophysiology}, 128(10):1994--2005, 2017.

\bibitem{dericioglu2018success}
Nese Dericioglu and P{\i}nar Ozdemir.
\newblock The success rate of neurology residents in eeg interpretation after
  formal training.
\newblock {\em Clinical EEG and neuroscience}, 49(2):136--140, 2018.

\bibitem{halford2017characteristics}
Jonathan~J Halford, Amir Arain, Giridhar~P Kalamangalam, Suzette~M LaRoche,
  Bonilha Leonardo, Maysaa Basha, Nabil~J Azar, Ekrem Kutluay, Gabriel~U Martz,
  Wolf~J Bethany, et~al.
\newblock Characteristics of eeg interpreters associated with higher interrater
  agreement.
\newblock {\em Journal of clinical neurophysiology: official publication of the
  American Electroencephalographic Society}, 34(2):168, 2017.

\bibitem{lhatoo2005cause}
Samden~D Lhatoo and Josemir~WAS Sander.
\newblock Cause-specific mortality in epilepsy.
\newblock {\em Epilepsia}, 46:36--39, 2005.

\bibitem{halford2018interictal}
Jonathan~J Halford, M~Brandon Westover, Suzette~M LaRoche, Micheal~P Macken,
  Ekrem Kutluay, Jonathan~C Edwards, Leonardo Bonilha, Giridhar~P Kalamangalam,
  Kan Ding, Jennifer~L Hopp, et~al.
\newblock Interictal epileptiform discharge detection in eeg in different
  practice settings.
\newblock {\em Journal of Clinical Neurophysiology}, 35(5):375--380, 2018.

\bibitem{santoshkumar2009prevalence}
Balagopal Santoshkumar, Jaron~JR Chong, Warren~T Blume, Richard~S McLachlan,
  G~Bryan Young, David~C Diosy, Jorge~G Burneo, and Seyed~M Mirsattari.
\newblock Prevalence of benign epileptiform variants.
\newblock {\em Clinical Neurophysiology}, 120(5):856--861, 2009.

\bibitem{halford2009computerized}
Jonathan~J Halford.
\newblock Computerized epileptiform transient detection in the scalp
  electroencephalogram: Obstacles to progress and the example of computerized
  ecg interpretation.
\newblock {\em Clinical Neurophysiology}, 120(11):1909--1915, 2009.

\bibitem{scheuer2017spike}
Mark~L Scheuer, Anto Bagic, and Scott~B Wilson.
\newblock Spike detection: Inter-reader agreement and a statistical turing test
  on a large data set.
\newblock {\em Clinical Neurophysiology}, 128(1):243--250, 2017.

\bibitem{halford2013standardized}
Jonathan~J Halford, Robert~J Schalkoff, Jing Zhou, Selim~R Benbadis, William~O
  Tatum, Robert~P Turner, Saurabh~R Sinha, Nathan~B Fountain, Amir Arain,
  Paul~B Pritchard, et~al.
\newblock Standardized database development for eeg epileptiform transient
  detection: Eegnet scoring system and machine learning analysis.
\newblock {\em Journal of neuroscience methods}, 212(2):308--316, 2013.

\bibitem{spikeDetectionReview2002}
Scott~B Wilson and Ronald Emerson.
\newblock Spike detection: a review and comparison of algorithms.
\newblock {\em Clinical Neurophysiology}, 113(12):1873--1881, 2002.

\bibitem{spikeDetectionReview2018}
Fathi E.~Abd El{-}Samie, Turky~N. Alotaiby, Muhammad~Imran Khalid, Saleh~A.
  Alshebeili, and Saeed~Abdullah Aldosari.
\newblock A review of {EEG} and {MEG} epileptic spike detection algorithms.
\newblock {\em {IEEE} Access}, 6, 2018.

\bibitem{DawidSnake1979}
Alexander~Philip Dawid and Allan~M Skene.
\newblock Maximum likelihood estimation of observer error-ratesusing the em
  algorithm.
\newblock {\em Applied statistics}, pages 20--28, 1979.

\bibitem{Zhang2014}
Denny~Zhou Yuchen~Zhang, Xi~Chen and Michael~I Jordan.
\newblock Spectral methods meet em: A provablyoptimal algorithm for
  crowdsourcing.
\newblock {\em NIPS}, pages 1260–--1268, 2014.

\bibitem{rodrigues2014gaussian}
Filipe Rodrigues, Francisco Pereira, and Bernardete Ribeiro.
\newblock Gaussian process classification and active learning with multiple
  annotators.
\newblock In {\em International Conference on Machine Learning}, pages
  433--441, 2014.

\bibitem{BraggWeld2016}
et~al. Jonathan~Bragg, Daniel S~Weld.
\newblock Optimal testing for crowd workers.
\newblock {\em Proceedings of the 2016 International Conference on Autonomous
  Agents and Multiagent Systems}, pages 966–--974, 2016.

\bibitem{Khetan2018}
Zachary C.~Lipton Ashish~Khetan and Anima Anandkumar.
\newblock Learning from noisy singly-labeled data.
\newblock {\em arXiv prepring}, 2018.

\bibitem{Gupta2016ModelingMT}
Rahul Gupta, Kartik Audhkhasi, Zach Jacokes, Agata Rozga, and Shrikanth
  Narayanan.
\newblock Modeling multiple time series annotations as noisy distortions of the
  ground truth: An expectation-maximization approach.
\newblock {\em IEEE Transactions on Affective Computing}, 9:76--89, 2016.

\bibitem{Wang2018TowardsBetterGoldStandard}
Chen Wang, Phil Lopes, Thierry Pun, and Guillaume Chanel.
\newblock Towards a better gold standard: Denoising and modelling continuous
  emotion annotations based on feature agglomeration and outlier
  regularisation.
\newblock In {\em Proceedings of the 2018 on Audio/Visual Emotion Challenge and
  Workshop}, AVEC'18, pages 73--81, New York, NY, USA, 2018. ACM.

\bibitem{Natarajan2013}
Pradeep K~Ravikumar Nagarajan~Natarajan, Inderjit S~Dhillon and Ambuj Tewari.
\newblock Learning withnoisy labels.
\newblock {\em Advances in neural information processing systems}, pages
  1196--–1204, 2013.

\bibitem{Whitehill2009}
Jacob Bergsma Javier R~Movellan Jacob~Whitehill, Ting-fan~Wu and Paul~L Ruvolo.
\newblock Whose vote should count more: Optimal integration of labels from
  labelers of unknown expertise.
\newblock {\em Advances in neural information processing systems}, pages
  2035–--2043, 2009.

\bibitem{Khetan2016}
Ashish Khetan and Sewoong Oh.
\newblock Achieving budget-optimality with adaptive schemes in crowd-sourcing.
\newblock {\em Advances in Neural Information Processing Systems}, pages
  4844–--4852, 2016.

\bibitem{Sheng2008}
Foster~Provost Victor S~Sheng and Panagiotis~G Ipeirotis.
\newblock Get another label? improving data quality and data mining using
  multiple, noisy labelers.
\newblock {\em ACM SIGKDD international conference on Knowledge discovery and
  data mining}, pages 614–--622, 2008.

\bibitem{Ipeirotis2014}
Victor S~Sheng Panagiotis G~Ipeirotis, Foster~Provost and Jing Wang.
\newblock Repeated labeling using multiple noisy labelers.
\newblock {\em Data Mining and Knowledge Discovery}, pages 402--441, 2014.

\bibitem{guan2018said}
Melody~Y Guan, Varun Gulshan, Andrew~M Dai, and Geoffrey~E Hinton.
\newblock Who said what: Modeling individual labelers improves classification.
\newblock In {\em AAAI Conference on Artificial Intelligence}, 2018.

\bibitem{Rodrigues2018crowds}
Filipe Rodrigues and Francisco Pereira.
\newblock Deep learning from crowds.
\newblock {\em AAAI Conference on Artificial Intelligence}, 2018.

\bibitem{Huang2018crowdLayer}
Jian Huang, Ya~Li, Jianhua Tao, Zheng Lian, Mingyue Niu, and Minghao Yang.
\newblock Deep learning for continuous multiple time series annotations.
\newblock In {\em Proceedings of the 2018 on Audio/Visual Emotion Challenge and
  Workshop}, AVEC'18, pages 91--98, New York, NY, USA, 2018. ACM.

\end{thebibliography}

\end{document}